\documentclass{article}

\PassOptionsToPackage{numbers, compress}{natbib}

 \usepackage[dblblindworkshop, final]{neurips_2025}
\workshoptitle{Learning from Time-Series for Health}

\usepackage[utf8]{inputenc} 
\usepackage[T1]{fontenc}    
\usepackage{hyperref}       
\usepackage{url}            
\usepackage{booktabs}       
\usepackage{amsfonts}       
\usepackage{nicefrac}       
\usepackage{microtype}      
\usepackage{xcolor}         

\usepackage{bm} 
\usepackage{graphicx}
\usepackage{amsmath} 
\usepackage{amssymb}  

\DeclareMathOperator*{\argmin}{arg\,min}

\title{Convolutional Monge Mapping between EEG Datasets to Support Independent Component Labeling}

\author{%
  Austin Meek \\
  Department of Computer and Information Sciences\\
  University of Delaware\\
  Newark, Delaware \\
  \texttt{ajmeek@udel.edu} \\
  \And
  Carlos H. Mendoza-Cardenas \\
  Twitch Interactive Inc. \\
  San Francisco, California \\
  \AND
  Austin J. Brockmeier \\
  Department of Electrical and Computer Engineering \\
  Department of Computer and Information Sciences \\
  University of Delaware \\
  Newark, Delaware \\
}

\begin{document}

\maketitle

\begin{abstract}
EEG recordings contain rich information about neural activity and are used in the diagnosis and monitoring of multiple neuropathologies, including epilepsy and psychosis, but are subject to artifacts and noise.  While EEG analysis can benefit from automating artifact removal through independent component analysis and automatic labeling of independent components (ICs), differences in recording equipment and context (the presence of noise from electrical wiring and other devices) may impact the performance of IC classifiers. 
Here we investigate how these differences can be minimized by appropriate spectral normalization through filtering using Convolutional Monge Mapping Normalization (CMMN), which was previously shown to improve deep neural network approaches for sleep staging. We propose a novel extension of the CMMN method with two alternative approaches to computing the source reference spectrum the target signals are mapped to: (1) channel-averaged and $l_1$-normalized barycenter, and (2) a subject-to-subject mapping that finds the source subject with the closest spectrum to the target subject. Notably, our extension yields space-time separable filters that can be used to map between datasets with different numbers of EEG channels. We apply these filters in an IC classification task, and show significant improvement in recognizing brain versus non-brain ICs.
\end{abstract}




 In the context of EEG, differences in set-up (electrodes, amplifiers, analog and digital filters, power line noise)  can create stark changes in the spectral content of recordings, which can deteriorate the performance of models, especially when trained on limited data. A solution to this problem is the Convolutional Monge Mapping Normalization (CMMN) approach, a method for spectral normalization based on optimal transport~\citep{gnassounouConvolutionMongeMapping2023}. The CMMN method optimizes a unique linear filter for each subject and channel. After filtering, the spectra for each channel across all subjects are aligned to a common spectrum, which is the barycenter of the training subjects' spectra for that channel. Originally, CMMN was applied to sleep staging with deep neural networks. In this work, we extend it to the case of automatically labeling independent components, such as the popular ICLabel classifier~\citep{Pion-Tonachini2019}, which can support automatic EEG artifact removal and denoising. In comparison to the ICLabel classifier, which uses a neural network model with scalp map and spectral features (including the autocorrelation sequence), and was trained on an extensive dataset~\citep{pion2019iclabel}, we consider custom classifiers trained on a much smaller dataset collected in the US~\citep{onton2009high} using only time-series features extracted from each independent component~\citep{mendoza-cardenasLabelingEEGComponents2023}—that is, without any spatial information. We then test on an another independent  dataset~\citep{gramann2010}, collected in Europe. These two datasets were used in a prior work on IC classification~\citep{frolichClassificationIndependentComponents2015}, and are now both available on OpenNeuro~\citep{Onton2022,ds006563}. Because these datasets have expert-labeled ICs, we avoid filtering each channel with a different filter, which would change each IC differently; instead, we find a single CMMN filter for each subject that is applied to all channels. This channel-averaged CMMN filter is based on matching the channel-averaged spectrum of each test subject to either the barycenter of the channel-averaged spectra from the training set or the nearest spectrum in the training set in terms of the optimal transport distance. With a shared single temporal filter across channels, the filtering can be applied before or after the spatial filtering achieved by ICA's demixing matrix. Crucially, the use of the channel-averaging  enables the application of CMMN to target domains with a differing number of channels. Here, the source (train) and target (test) datasets have 134–235 channels and 64 channels, respectively. This extension of the CMMN methodology achieves domain adaptation between two datasets with the pre-trained classifiers such that their performance (the subject averaged F1 score for the brain class is 0.91) exceeds the performance of ICLabel (0.88).

\section{Methodology}
The CMMN methodology enables domain adaptation between different neural signals \citep{gnassounouConvolutionMongeMapping2023} by mapping the spectra of the target subjects to the barycenter of the source subjects via  a linear filter that achieves the optimal transport with respect to an $\ell_2$-metric, assuming the signals are zero-mean stationary Gaussian discrete-time processes (see Appendix~\ref{app} for more details).  
We extend this methodology in two key ways: we propose to use channel-averaged PSDs yielding a single filter per subject and a subject-to-subject mapping scheme. A key benefit of using a single filter per subject is that it enables adapting between EEG datasets with a variable number of channels.  We also propose to use $\ell_1$-normalization of spectra for invariance to signal scale, which may occur due to impedance differences due to electrodes or scalp contact or differences in the electronic systems (amplifiers and filtering) or digital filtering.



We assume access to  data (multichannel EEG) from $I$ subjects in the source domain and one or more target subjects. We apply CMMN to transform the target EEGs to the source domain and apply classifiers trained on the source domain to the transformed target data. Let $x_c[n]$ denote the $c$-th channel of the target signal, $c\in\{1,\ldots,C^\text{T}\}$. The CMMN transformation is a linear filtering of $x_c[n]$, with output $y_c[n] = (h * x_c)[n], \quad c \in \{1,\ldots, C^\text{T}\}$, where $h[n]$ is the normalizing CMMN filter, and $y_c[n]$ is the transformed signal that will have a source-like spectrum (the channel-averaged PSDs will match). The estimation of the normalizing CMMN filter is a three-step process that uses a reference spectrum (computed in the second step) that is either a \textbf{Barycenter} from the source or the normalized PSD of a source subject via the subject-to-subject (\textbf{Subj-to-subj}) assignment: 1) power spectral density calculation, 2a) barycenter calculation, 2b) minimization of the Hellinger distance between target signals and source signals for the \textbf{Subj-to-subj} mapping scheme, and 3) calculation of the normalizing filters.

\textbf{Step 1}  PSD  calculation. As in \citep{gnassounouConvolutionMongeMapping2023}, we use the Welch periodogram method \citep{welchUseFastFourier1967}, which takes averages of the squared FFT from possibly overlapping windows, to calculate PSDs for each subject, as implemented by the SciPy library \citep{2020SciPy-NMeth}.  For real-valued signals, the spectra will be conjugate symmetric for positive and negative frequencies. In this case, $\bm{p}\in\mathbb{R}_{\ge 0}^{P}$ denotes the PSD for $P$ non-negative frequencies (for zero-mean signals the DC component should be zero but is retained). For $K$ equal-sized windows of length $M$, $\{\bm{x}_k \}_{k=1}^K=\mathcal{X}\subset \mathbb{R}^M $ with windowing function $\bm{w}$, the power spectral density estimate is $\bm{p}= \frac{1}{|\mathcal{X}|}\sum_{\bm{x}\in\mathcal{X}} \lvert \mathrm{RFFT}(\bm{w}\odot \bm{x})\rvert^{\odot 2}\in \mathbb{R}_{\ge0}^{P}$, where $\mathrm{RFFT}$ denotes the FFT operation for real-valued signals and returns the density estimate only for non-negative frequencies, $P = \mathtt{nfft}/2+1$ is the number of non-negative frequencies, and $N=\mathtt{nfft}$ is the length of the windows after zero-padding, assumed to be even. For a subject with $C$ channels, the matrix of PSDs is $\bm{P}=[\bm{p}_1,\ldots,\bm{p}_C]^\top \in \mathbb{R}^{C\times P}$. 
Unlike the original CMMN, which used different filters for each channel, we propose to compute a channel-averaged PSD as
$\bar{\bm{p}} = \frac{1}{C_i} \sum_{c=1}^{C} \bm{p}_{c}$.\footnote{One possible concern is whether channel-averaging is a sufficient characterization of multi-channel EEG given the spectral differences in signals across the scalp. Of course, if the location of the electrodes on the scalp is vastly different such that the spectral content is not comparable, then channel-averaged PSD may not be meaningful, but for EEG montages with whole scalp this should not be an issue.}

\textbf{Step 2a} Barycenter calculation. We compute a barycenter for the source subjects only, this serves as the template for mapping the target subjects. For $I$ source subjects with channel-averaged PSDs $\bar{\bm{p}}_1^\text{S},\ldots,\bar{\bm{p}}_I^\text{S}$, the  barycenter, $\bar{\bm{p}}_\text{S}$, and the $\ell_1$-normalized barycenter, $\tilde{\bm{p}}_\text{S}$, are defined as
\begin{align} \label{eq:2:barycenter_1}
    {\bar{\bm{p}}}_\text{S} = \frac{1}{I} \sum_{i=1}^{I} \bar{\bm{p}}^\text{S}_{i}, \quad  {\tilde{\bm{p}}}_\text{S} = \frac{1}{I} \sum_{i=1}^{I} \frac{\bar{\bm{p}}^\text{S}_i}{\|\bar{\bm{p}}^\text{S}_i\|_1}=\frac{1}{I} \sum_{i=1}^{I} \tilde{\bm{p}}^\text{S}_i,\quad  \tilde{\bm{p}}^\text{S}_i= \frac{\bar{\bm{p}}^\text{S}_i}{\|\bar{\bm{p}}^\text{S}_i\|_1},
\end{align}
The $l_1$ normalization ensures that each subject contributes equally to the average. As the units of PSD are squared, outliers without normalization can influence the average.

\textbf{Step 2b: Subject-to-Subject Mapping.} As an alternative to the barycenter scheme, we map a target subject to the nearest source subject. For a target subject with an $\ell_1$-normalized PSD $\tilde{\bm{p}}^\text{T}= \frac{\bar{\bm{p}}^\text{T}}{\|\bar{\bm{p}}^\text{T}\|_1}$, we find the index $i^*$ of the source subject that minimizes the Hellinger distance, which is applicable since the $\ell_1$-normalized PSD is a probability mass function (PMF)—non-negative and sums to 1. The channel-averaged PSD of the best-matched source is then $\bar{\bm{p}}_*^\text{S} = \bar{\bm{p}}_{i^*}^\text{S}$, where
\begin{align}
    \label{eq:best_psd}
    i^* = \argmin_{i\in\{1,\ldots,I\}}  d_\text{He}( \tilde{\bm{p}}^{\text{S}}_i,\tilde{\bm{p}}^{\text{T}}) =\frac{1}{\sqrt{2}}\left\lVert (\tilde{\bm{p}}^{\text{S}}_i)^{\odot\frac{1}{2}} -(\tilde{\bm{p}}^{\text{T}})^{\odot\frac{1}{2}} \right\rVert_2  =\sqrt{\frac{1}{2} \sum_{n=0}^{N-1} \left(\sqrt{\tilde{p}_i^\text{S}[n]} - \sqrt{\tilde{p}^\text{T}[n]}\right) ^2},
\end{align}
where  $d_\text{He}$ is the Hellinger distance.  Assuming that $\tilde{\bm{p}}^{\text{S}}_i$, $\tilde{\bm{p}}^{\text{T}}$ are the eigenvalues of the trace normalized circulant matrices $\bm{\rho}^\text{S}_i =\frac{\bm{\Sigma}^\text{S}_i}{\mathrm{tr}(\bm{\Sigma}^\text{S}_i) } $ and $\bm{\rho}^\text{T} =\frac{\bm{\Sigma}^\text{T}}{\mathrm{tr}(\bm{\Sigma}^\text{T})}$, where $\bm{\Sigma}^\text{S}_i$ and $\bm{\Sigma}^\text{T}$ are the covariance matrices of the discrete-time  zero-mean stationary Gaussian random processes, then the Wasserstein-2 distance between the two variance-normalized processes is
 \begin{align}
\label{eq:hellinger}
     W_2(&\mathcal{N}(\bm{0},\bm{\rho}^\text{S}_i), \mathcal{N}(\bm{0},\bm{\rho}^\text{T})) = \sqrt{2}   d_\text{He}( \tilde{\bm{p}}^{\text{S}}_i,\tilde{\bm{p}}^{\text{T}}).
\end{align}
The use of Hellinger distance compared to Bures-Wasserstein distance means that signals are compared in terms of their spectral shape, ignoring difference in their variance.  

\textbf{Step 3} Normalizing filter. We define a real-valued, zero-phase linear filter with impulse response  $h[n]$ and frequency response $\bm{H}$ given by the square root of the ratio of the channel-averaged PSDs~\citep{flamaryConcentrationBoundsLinear2020}:
\begin{align} \label{eq:6:ot_mapping}
    \bm{h} &= \mathrm{IRFFT}_M(\bm{H}),\quad  
    \bm{H} = \bar{\bm{p}}^{\text{T} \odot -\frac{1}{2}} \odot \bar{\bm{p}}^{\text{S} \odot \frac{1}{2}},\quad 
     H[n]  = \frac{\sqrt{\bar{p}^\text{S}[n]}}{\sqrt{\bar{p}^\text{T}[n]}}, n\in\{0,\ldots, P-1\},
\end{align}
where $\mathrm{IRFFT}_M$ is the  inverse FFT for real-valued signals that operates on the non-negative frequencies. As discussed in Appendix~\ref{app}, by equalizing the channel-average PSD, this filter solves the optimal transport problem between zero-mean Gaussian distributions with circulant covariance matrices that commute and are diagonalized by the discrete-time Fourier transform. The only difference between the barycenter and subject-to-subject formulations is the choice of Gaussian distribution for the source. That is, the source PSD $\bar{\bm{p}}^\text{S}$ can be either the barycenter of the source dataset $\bar{\bm{p}}_\text{S}$ as in \eqref{eq:2:barycenter_1}, the $\ell_1$-normalized barycenter  $\tilde{\bm{p}}_\text{S}$  as in \eqref{eq:2:barycenter_1}, or the closest matching source signal $\bar{\bm{p}}_*^\text{S}$ as in \eqref{eq:best_psd}.

\subsection{Application of CMMN to pre-trained IC classifiers}
While the proposed methodology is general and could be applied to normalize any target dataset for classifiers pre-trained (without CMMN) on a source dataset, we focus on classifiers for automatically labeling independent components of EEG into brain and other classes~\citep{winkler2011automatic,frolichClassificationIndependentComponents2015,Pion-Tonachini2019}. In particular, we build on recent work for labeling ICs without spatial information~\citep{mendoza-cardenasLabelingEEGComponents2023}, applying the CMMN approach to random forest classifiers pre-trained on two different feature vectors: the PSD \footnote{In contrast to the original ICLabel implementation, we use a min-max normalization of the log-scale PSD instead of dividing by its max-absolute value because it is more robust to the scale of the EEG signals and provided a better classification performance.} and autocorrelation sequence of the independent components, as in the MNE-ICALabel implementation~\citep{Li2022}, a Python port of ICLabel~\citep{Pion-Tonachini2019}.

\section{Experiments and Results}
We use two datasets for this domain adaptation study, the \textit{Imagined Emotion} (\textit{Emotion} for short) dataset from \citep{Onton2022} as the source dataset and the \textit{Cue} dataset from \citep{gramann2010} as the target dataset. Both datasets include manually labeled independent components (ICs). The \textit{Emotion} dataset has data from 32 subjects (13 male and 19 female, with an age mean and standard deviation of 25.5 $\pm$ 5 years). The EEG data has 180-232 channels and sampled at 256 Hz with durations ranging from 54 to 136 minutes. The dataset includes 935 expert-labeled ICs in three categories: brain (570), muscle (306), and eye (59). We used 27 subjects for training and 7 for testing. The training set contains 5,786 total ICs. The \textit{Cue} dataset has data from 12 subjects (10 male, 2 female, with an age range of 21 to 25 years), using 64 channels during data collection. This dataset's recordings were 56 to 66 minutes long and were collected at 500 Hz, which we down-sampled to 256 Hz before performing domain adaptation.  Expert annotations are available for 389 ICs: brain (261), muscle (102), eye (22), and heart (4). This dataset serves as the target for cross-dataset evaluation, testing generalization of the classifiers trained on the \textit{Emotion}  training data. Note that in comparison to the experiments in the original CMMN paper \citep{gnassounouConvolutionMongeMapping2023}, our data contains different numbers of channels (134–235 vs. 64). Therefore, the use of channel-averaged PSDs for CMMN is essential. For further details on both datasets, please refer to \citep{Onton2022, gramann2010}.

\begin{figure}[htb]
    \centering
    \includegraphics[width=0.495\linewidth]{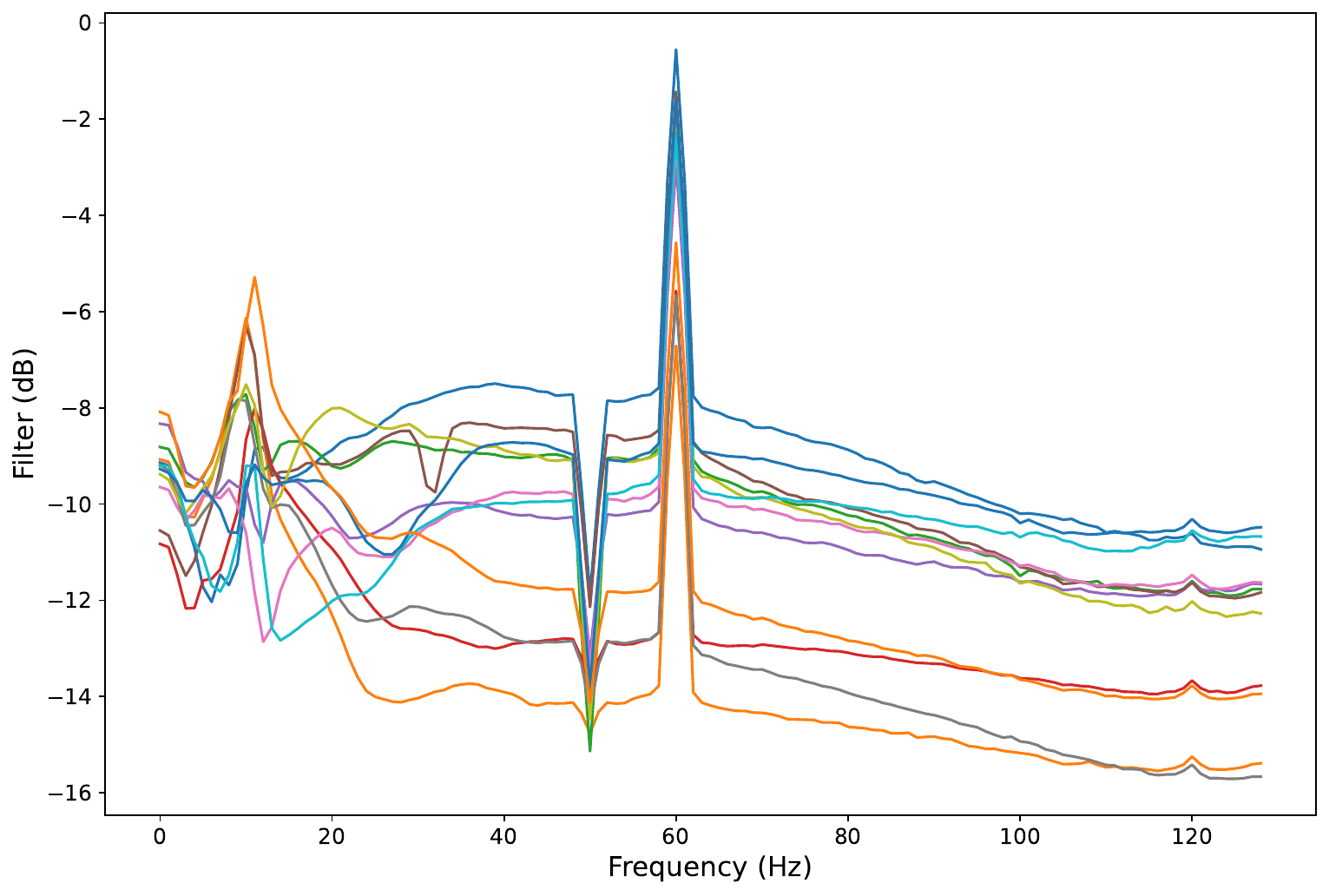}
    \includegraphics[width=0.495\linewidth]{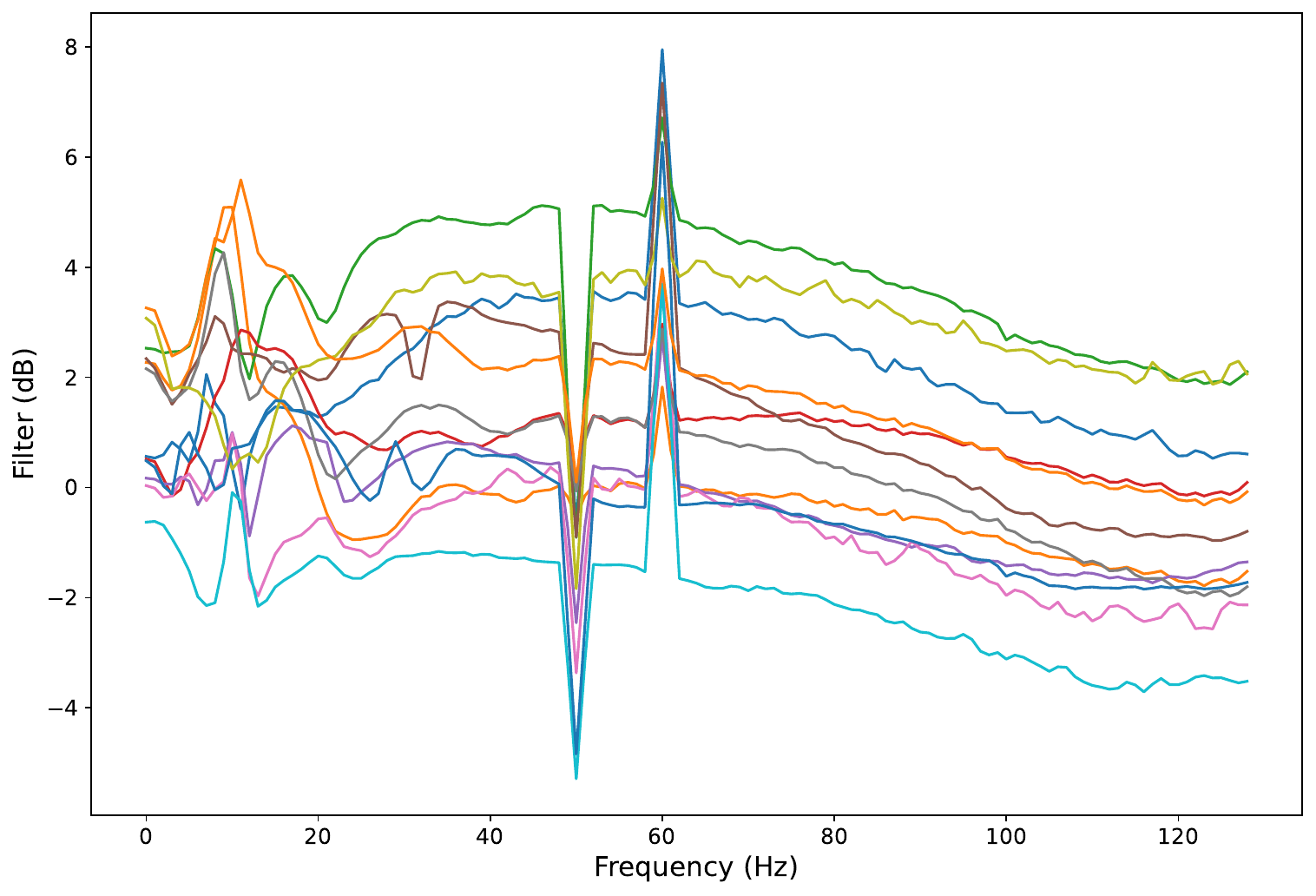}  
    \caption{Frequency response of filters learned for mapping the \textit{Cue} dataset to the \textit{Emotion} dataset. (Left) \textbf{Barycenter} mapping scheme with a $\ell_1$-normalized barycenter. Notice that the noise at 50 Hz is reduced and the noise at 60 Hz is relatively amplified. Overall, the filter attenuates since the normalized barycenter has lower magnitudes.(Right) \textbf{Subj-to-subj} mapping scheme. Even though the mapping scheme is different, the line noises are visibly still being `swapped'.}
    \label{fig:cue_freq_filter}
\end{figure}

We show that the normalizing filters learned display intuitive mapping behaviors. The \textit{Emotion} dataset \citep{Onton2022} was collected in the US, and displays a characteristic 60 Hz line noise as seen in Fig.~\ref{fig:all_emotion_subj_psd} in the appendix. The \textit{Cue} dataset \citep{gramann2010} was collected in Europe, and displays a characteristic 50 Hz line noise as seen  in Fig.~\ref{fig:cue_subj_psd_raw} in the appendix. When mapping from the Europe dataset to the US dataset. We can see in Fig.~\ref{fig:cue_freq_filter} that the filters learned to minimize the 50 Hz spike and introduce a spike at 60 Hz. 

We trained multi-class classifiers using both PSD and Autocorrelation features on the subset of 27 subjects from \textit{Emotion} dataset, with hyperparameters selected by leave-one-subject-out (LOO) cross-validation. The classifier is trained on segments of length $l_\text{train}$ (a hyperparameter), and validated (using the F1 score on the brain class) on segments of length $l_\text{val} \in \{5, 50\}$ minutes, with $l_\text{train} \le l_\text{val}$. Given the best LOO hyperparameters, a new model is trained on the entire training set of 27 subjects. Here we report test results on segments of length equal to the validation segment length,  from a disjoint test set of 7 source \textit{Emotion} subjects and all the target subjects in the \textit{Cue} dataset.  We benchmark our method against ICLabel \citep{Pion-Tonachini2019}. As a model trained on an extensive dataset of expert- and crowd-labeled ICs, ICLabel serves as a strong baseline that can be applied to both the source and target datasets without our proposed CMMN. We did not apply ICLabel to the CMMN normalized signals,  as it was trained over a wide variety of datasets and should be already invariant.
The source code for our implementation and experiments is publicly available at \href{https://github.com/cniel-ud/ICWaves}{https://github.com/cniel-ud/ICWaves}. 



The performance on held-out subjects of the source \textit{Emotion} dataset is shown in Table~\ref{tab:emotion}. PSD/Autocorr has the best performance at both time lengths. 
\begin{table}[htb]
  \caption{F1 score of brain-labeled independent components of classifiers within domain on held-out subjects.}
    \label{tab:emotion}
    \centering
    \begin{tabular}{l l l}
    \hline 
    Classifier  & 5 minutes & 50 minutes\\
    \hline
         PSD/Autocorr.& \textbf{0.93}$\pm$0.05   & \textbf{0.96}$\pm$0.05\\
ICLabel &   0.88$\pm$0.05  & 0.89$\pm$0.07\\
\hline 
    \end{tabular}  
\end{table}


\begin{table*}[h]
\caption{F1 score of brain-labeled independent components  for different test segment lengths  and CMMN schemes. Best performance for a classifier at a segment length is underlined and best classifier for a segment length is bolded. The last column contains the p-value for a one-sided Wilcoxon rank-sign test between no filtering and CMMN filtered (normalized barycenter for PSD/Autocorr.).}
\label{tabular_data}
\begin{center}
\begin{tabular}{l l r r r r r }
\hline
Classifier & Len. & No filtering & Barycenter & $\ell_1$-norm. Bary.  & Subj-to-subj & p-value  \\
\hline
PSD/Autocorr.& 5  &0.77 $\pm$ 0.09 & 0.78$\pm$0.12 &  \underline{0.84} $\pm$ 0.07 & 0.79 $\pm$ 0.17  & 0.0046 \\
ICLabel& 5  & \textbf{0.88} $\pm$ 0.06 & &  &  &  \\
\hline 
PSD/Autocorr.& 50 & 0.83 $\pm$ 0.09&  0.86$\pm$0.09 & \underline{0.86} $\pm$ 0.08 & 0.85 $\pm$ 0.17 & 0.1696 \\
 ICLabel& 50 & \textbf{0.89} $\pm$ 0.05 & & &  &  \\
 \hline 
\end{tabular}
\end{center}
\end{table*}
As seen in Table~\ref{tabular_data}, our results show that, when compared to no filtering, using the appropriate CMMN scheme improves the performance of  \textit{Emotion}-trained classifiers on the \textit{Cue} dataset. 
The subject-to-subject mapping scheme improves the brain class F1-score from 0.77 to 0.79 at 5 minutes and from 0.83 to 0.85 at 50 minutes. With the $\ell_1$-normalized barycenter, the PSD/Autocorr classifiers' F1-score improved from 0.77 to 0.84 at 5 minutes, and from 0.83 to 0.86 at 50 minutes. We use a Wilcoxon rank sign test ($n=12$, one-sided alternative) to test whether the filtering (normalized barycenter) has a significant improvement in F1 score for the brain-labeled ICs, and find significant improvement at a level of 0.05 for the 5 minute length. 


\section{Conclusion}

In this work we made key extensions to the recently introduced CMMN methodology~\citep{gnassounouConvolutionMongeMapping2023} to enable domain adaptation between EEG datasets for independent component classification for artifact removal.  We introduce filters defined by channel-averaged PSDs along with a subject-to-subject mapping scheme, and show that our CMMN method results in improvements when classifying brain versus non-brain independent components, achieving domain adaptation between two different datasets with significant differences. 
Our method advances the work on EEG artifact-removal across distinct datasets, an area that is crucial for increasing the clinical utility of EEG recordings.

 \begin{ack}

This work was supported in part by the University of Delaware General University Research fund. This research was supported in part through the use of DARWIN computing system: DARWIN – A Resource for Computational and Data-intensive Research at the University of Delaware and in the Delaware Region, which is supported by NSF under Grant Number: 1919839, Rudolf Eigenmann, Benjamin E. Bagozzi, Arthi Jayaraman, William Totten, and Cathy H. Wu, University of Delaware, 2021, URL: https://udspace.udel.edu/handle/19716/29071.

The authors thank Laura Fr\o{}lich, Tobias Andersen, Klaus Gramann, and Morten M\o{}rup for graciously providing access to the \textit{Cue} dataset used for our experiments, and Prof. Gramann for allowing us to post it on OpenNeuro.

 \end{ack}

\bibliographystyle{plainnat} 

\bibliography{TBME24/NER23}

\appendix
\section{Background on Convolutional Monge Mapping Normalization}
\label{app}
The Wasserstein-2 distance between Gaussian distributions has a closed form~\citep{dowson1982frechet,gelbrich1990formula} known as the Bures-Wasserstein distance or Fréchet distance, expressed in terms of the means $\bm{m}^\text{S},\bm{m}^\text{T}$ and covariance matrices  $\bm{\Sigma}^\text{S},\bm{\Sigma}^\text{T}$, as
\begin{align}
    W_2(\mathcal{N}&(\bm{m}^\text{S},\bm{\Sigma}^\text{S}), \mathcal{N}(\bm{m}^\text{T},\bm{\Sigma}^\text{T}))
    \\\notag&=\sqrt{\lVert \bm{m}^\text{S}\!-\!\bm{m}^\text{T} \rVert_2^2\!+\mathrm{tr}(\bm{\Sigma}^\text{S}\!+\!\bm{\Sigma}^\text{T}\!-\!(\bm{\Sigma}^{\text{S}\frac{1}{2}}\bm{\Sigma}^\text{T} \bm{\Sigma}^{\text{S}\frac{1}{2}} )^\frac{1}{2} )}, 
\end{align}
where the second term in the square root is the squared Bures distance. Under the stationarity assumption, a discrete-time zero-mean Gaussian process is completely described by its auto-covariance matrix, which is a symmetric and Toeplitz matrix formed from the auto-correlation sequence $r[k]=\mathbb{E}[x[n]x[n+k]]$. Assuming a sufficiently long truncation of $r[\tau]$ that yields a $N \times N$ circulant matrix that is positive-semidefinite, then the  discrete Fourier transform (DFT) matrix $\bm{F}$ diagonalizes it, $\bm{F} \bm{\Sigma}\bm{F}^H= \mathrm{diag}(\bm{\lambda})$, where $\bm{F}^H$ denotes the Hermitian transpose of $\bm{F}$. The resulting spectrum $
\bm{\lambda}=\mathrm{diag}(\bm{F} \bm{\Sigma}\bm{F}^H)$ corresponds to an statistical estimate of the power spectral density (PSD). The squared Wasserstein-2 distance between two discrete-time, zero-mean stationary Gaussian random processes with auto-covariance $\bm{\Sigma}^\text{S}$ and $\bm{\Sigma}^\text{T}$, which are circulant, simplifies to
\begin{equation}
\label{eq:w2_spectra}
     W_2(\mathcal{N}(\bm{0},\bm{\Sigma}^\text{S}), \mathcal{N}(\bm{0},\bm{\Sigma}^\text{T}))=\lVert \bm{\lambda}^{\text{S} \odot\frac{1}{2}} - \bm{\lambda}^{\text{T} \odot\frac{1}{2}}  \rVert_2=\sqrt{\sum_{n=0}^{N-1} \left(\sqrt{\lambda^\text{S}[n]} - \sqrt{\lambda^\text{T}[n]}\right) ^2}.
\end{equation}
The Bures-Wasserstein barycenter, which is the Gaussian special case of the Wasserstein-2 barycenter~\citep{agueh2011barycenters,bhatia2019bures}, is the covariance matrix $\bm{\Sigma}_\text{S}$ (in the set of positive semidefinite matrices $\mathcal{S}_N$) that minimizes the sum of squared Wasserstein-2 distances to a set of zero-mean Gaussian distributions described by covariance matrices $\{\bm{\Sigma}_i^\text{S}\}_{i=1}^I$:
\begin{align}
\bm{\Sigma}_\text{S}=    \argmin_{\bm{\Sigma} \in \mathcal{S}_N}  \sum_{i=1}^I \frac{1}{I} W_2^2(\mathcal{N}(\bm{0},\bm{\Sigma}), \mathcal{N}(\bm{0},\bm{\Sigma}_i^\text{S}),
\end{align}
Assuming all covariance matrices are circulant and positive semidefinite, it is straightforward to show that 
$\bm{\Sigma}_\text{S}= \bm{F}^H \mathrm{diag}(\bar{\bm{\lambda}}_\text{S})\bm{F}$, where $\bar{\bm{\lambda}}_\text{S}=(\frac{1}{I} \sum_i \bm{\lambda}_i^{\text{S}\odot \frac{1}{2}})^{\odot 2}$.

Consider a stationary target signal $x[n]$ with PSD $\bm{\lambda}^\text{T}=\mathbb{E}[|  \bm{F}\bm{x} |^{\odot 2} ]$ that is convolved with the filter impulse response $h[n]$ (frequency response $\bm{H}$), yielding the signal $y[n]=(h*x)[n]$. Then under the stationarity assumption, the PSD of $y[n]$ is 
\begin{align}
    \bm{\lambda}^y &= \mathbb{E}[|\bm{F} \bm{y} |^{\odot 2}]=\mathbb{E}[|  (\bm{F}\bm{x})\odot \bm{H} |^{\odot 2}]
    =\mathbb{E}[|  \bm{F}\bm{x} |^{\odot 2} \odot | \bm{H} |^{\odot 2}] = | \bm{H} |^{\odot 2} \odot \bm{\lambda}^\text{T},
\end{align} where $\bm{H}$ is the discrete Fourier transform of the filter. Assuming $x[n] $ is also zero-mean and a Gaussian process $\bm{x}\sim \mathcal{N}(\bm{0},\bm{\Sigma}^\text{T})$, then  $y[n]$ has a circulant auto-covariance matrix $\bm{\Sigma}^y = \bm{F} \mathrm{diag}( | \bm{H} |^{\odot 2} \odot \bm{\lambda}^\text{T} ) \bm{F}^H$. The Wasserstein-2 distance between a source process (possibly the barycenter) and the filtered target process is then 
\begin{equation} 
 W_2(\mathcal{N}(\bm{0},\bm{\Sigma}^\text{S}), \mathcal{N}(\bm{0},\bm{\Sigma}^y))=\lVert \bm{\lambda}^{\text{S} \odot\frac{1}{2}} - | \bm{H} | \odot \bm{\lambda}^{\text{T} \odot\frac{1}{2}}  \rVert_2^2.
\end{equation}
Clearly, the distance is zero for any filter $h[n]$ such that $
   | \bm{H}|=\bm{\lambda}^{\text{S} \odot\frac{1}{2}} \odot\bm{\lambda}^{\text{T} \odot-\frac{1}{2}}\in\mathbb{R}_{\ge 0}^N$, i.e., where the magnitude of the frequency response is the square-root of the power spectral densities $|H[n]|=\sqrt{ \frac { {\lambda}^{\text{S}}[n] } {\lambda^\text{T}[n]}}, n\in\{0,\ldots,N-1\}$. 

\newpage

\section{Additional Results}
\begin{figure}[htb]
    \centering
    \includegraphics[width=0.5\linewidth]{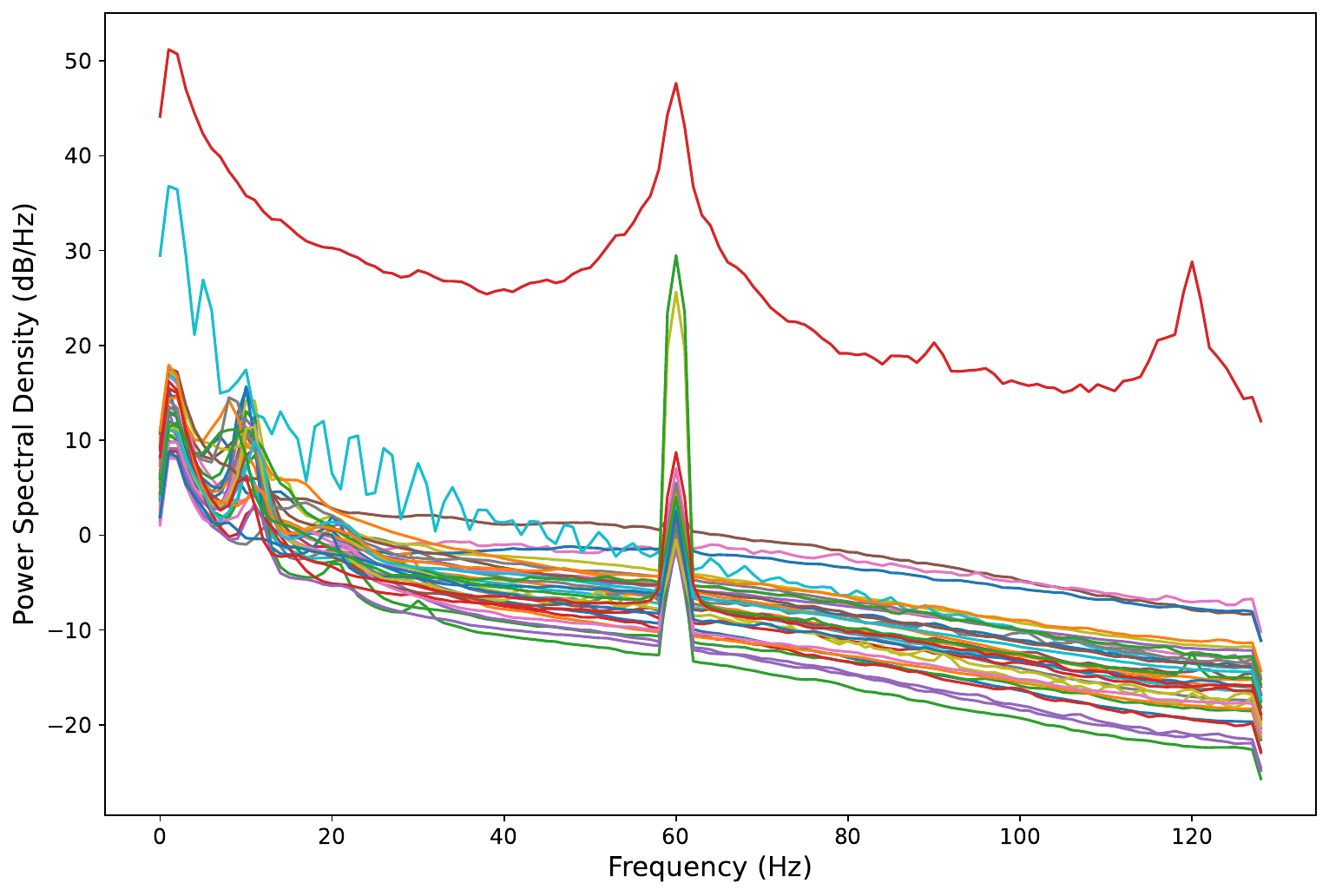}
    \includegraphics[width=0.5\linewidth]{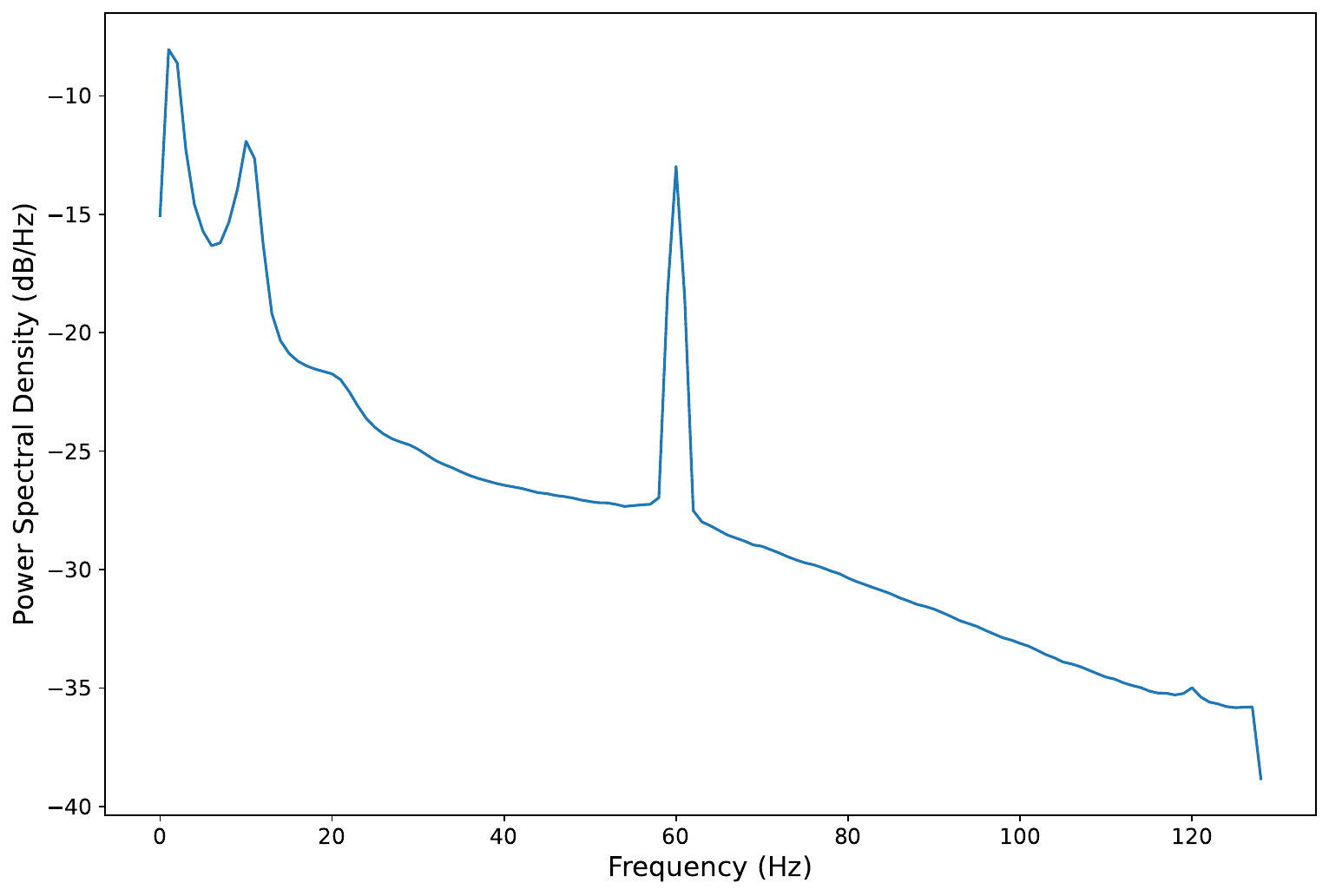}
    \caption{(Top) Channel-averaged power spectral densities for each subject in the \textit{Emotion}  dataset. Notice that some subjects are outliers in terms of overall amplitude. (Bottom) This is the $\ell_1$-normalized barycenter computed from the source \textit{Emotion} dataset. In the \textbf{Barycenter} mapping scheme, all target signals are filtered such that their robust channel-average PSD matches this. Notice the large spike at 60 Hz.}
    \label{fig:all_emotion_subj_psd}
\end{figure}

\begin{figure}[htb]
    \centering
    \includegraphics[width=0.5\linewidth]{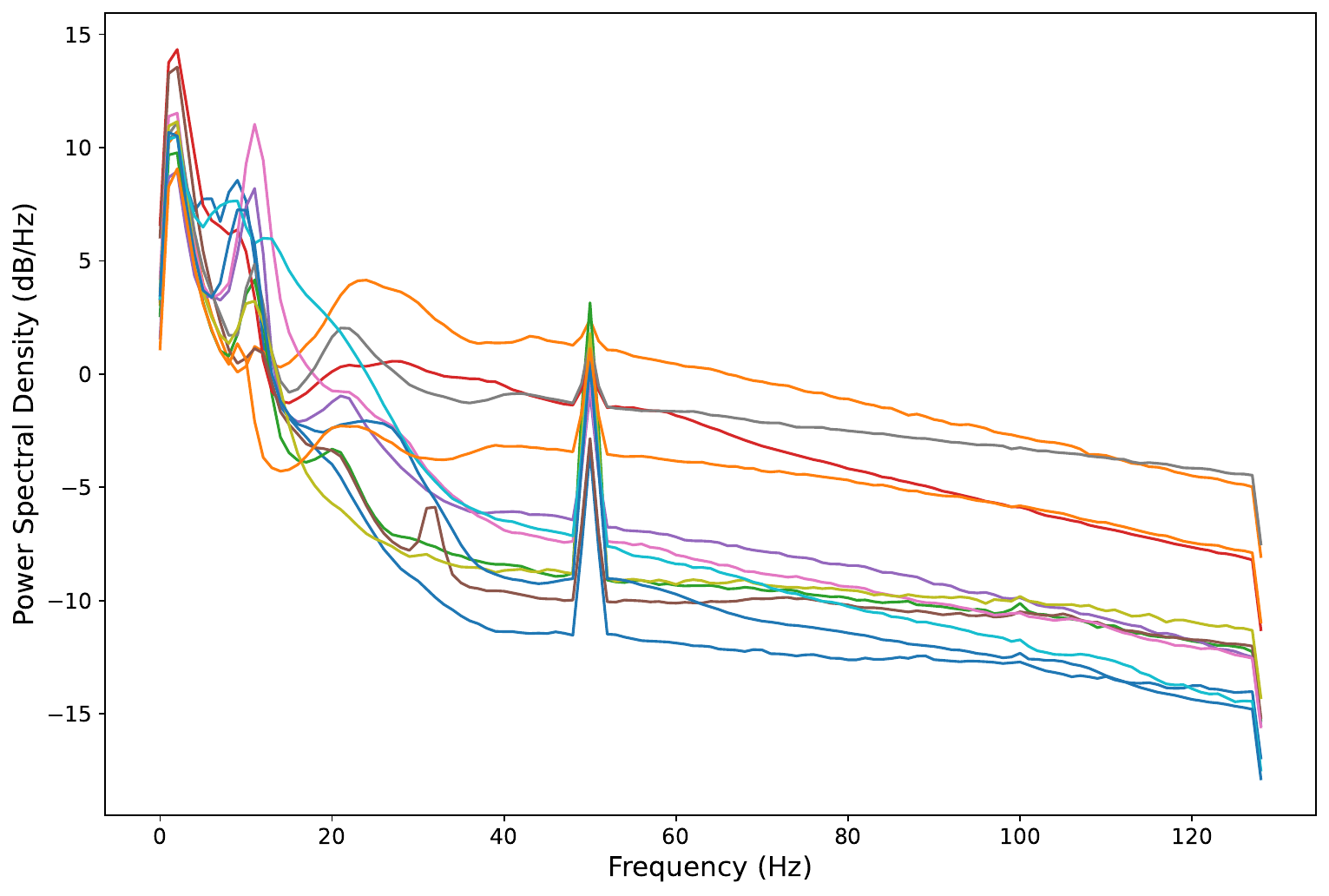}
    \caption{PSDs for different subjects in the target \textit{Cue} dataset. Notice the spike at 50 Hz.}
    \label{fig:cue_subj_psd_raw}
\end{figure}

\end{document}